\documentclass[10pt,twocolumn,letterpaper]{article}

\usepackage{cvpr}
\usepackage{times}
\usepackage{epsfig}
\usepackage{graphicx}
\usepackage{amsmath}
\usepackage{amssymb}
\usepackage[ruled,vlined,linesnumbered]{algorithm2e}
\usepackage[noend]{algpseudocode}
\usepackage{enumerate}
\usepackage{comment}
\usepackage{bbding}

\usepackage{xspace}

\usepackage[pagebackref=true,breaklinks=true,letterpaper=true,colorlinks,bookmarks=false]{hyperref}

\cvprfinalcopy 

\def\cvprPaperID{0366} 

\ifcvprfinal\fi
\begin{document}

\title{SCD: A Stacked Carton Dataset for Detection and Segmentation}

\author{Jinrong Yang\dag\textsuperscript{\rm 1}\quad Shengkai Wu\dag\textsuperscript{\rm 1}\quad Lijun Gou\textsuperscript{\rm 1}\quad Hangcheng Yu\textsuperscript{\rm 1}\quad \\  Chenxi Lin\textsuperscript{\rm 1}\quad Jiazhuo Wang\textsuperscript{\rm 1}\quad Minxuan Li\textsuperscript{\rm 2} \quad Xiaoping Li*\textsuperscript{\rm 1}\\\\
	\textsuperscript{\rm 1}State Key Laboratory of Digital Manufacturing Equipment and Technology\\
	Huazhong University of Science and Technology, China.\\
	\textsuperscript{\rm 2}Faculty of Arts and Science, Queen's University, Canada\\
\dag Equal contribution\quad\quad\quad\quad\quad *Corresponding Author \\
{\tt\small \{yangjinrong, ShengkaiWu\}@hust.edu.cn}}

\maketitle
\begin{abstract}


Carton detection is an important technique in the automatic logistics system and can be applied to many applications such as the stacking and unstacking of cartons, the unloading of cartons in the containers. However, there is no public large-scale carton dataset for the research community to train and evaluate the carton detection models up to now, which hinders the development of carton detection. In this paper, we present a large-scale carton dataset named Stacked Carton Dataset(SCD) with the goal of advancing the state-of-the-art in carton detection. Images are collected from the internet and several warehourses, and objects are labeled using per-instance segmentation for precise localization. There are totally 250,000 instance masks from 16,136 images. In addition, we design a carton detector based on RetinaNet by embedding Boundary Guided Supervision module(BGS) and Offset Prediction between Classification and Localization module(OPCL). OPCL alleviates the imbalance problem between classification and localization quality which boosts AP by $3.1\% \sim 4.7\%$ on SCD while BGS guides the detector to pay more attention to boundary information of cartons and decouple repeated carton textures. To demonstrate the generalization of OPCL to other datasets, we conduct extensive experiments on MS COCO and PASCAL VOC. The improvements of AP on MS COCO and PASCAL VOC are $1.8\% \sim 2.2\%$ and $3.4\% \sim 4.3\%$ respectively.
\end{abstract}

\begin{table*}[t]
\begin{center}
\begin{tabular}{c|c c c c c c c}
\hline
{Dataset} & {Images} & {Labels} & {All/Occlusion} & {Inner/Outer} &  {Carton Label} & {Total Instances} & {Average Instances}\\
 \hline
ImageNet\cite{deng2009imagenet} & 1388  & 1 & \XSolidBrush & \XSolidBrush & \CheckmarkBold & - & - \\
Open Image\cite{2020The} & -  & 1 & \XSolidBrush & \XSolidBrush & \XSolidBrush & - & -\\
 \hline
LSCD & 7735  & 4\&1 &\CheckmarkBold & \CheckmarkBold & \CheckmarkBold & 81870 & 10.58\\
OSCD & 8401 & 1 & \XSolidBrush & \XSolidBrush & \CheckmarkBold & 168748 & 20.09\\
 \hline
\end{tabular}
\\ \hspace*{\fill} \\
\caption{The comparison of Datasets including the category of carton. "Labels" represents the number of carton class. '$4\&1$' in the LSCD column represents that it supports both 4 labels (Carton-inner-all, Carton-inner-occlusion, Carton-outer-all and Carton-outer-occlusion) and 1 label (Carton). Only LSCD supports occlusion information and inner/outer information by giving related annotations. "Images" means the number of images containing carton. "-" means the corresponding dataset only includes the category of box where carton only appear in images but the exact labels are not given in Open Image while too much noisy for ImageNet to calculate the number of instances. "Average Instances" represents the average number of instances in each image.}
\label{tab:1}
\end{center}
\end{table*}

\section{Introduction}
\label{section:1}

Carton detection and segmentation is extremely important for automatic logistics transport, \ie, robots perform stacking and unstacking operations, and especially for achieving the goal of complete automation of logistics scenarios. Traditional methods of carton localization include manual identification, RFID-based(Radio Frequency Identification) carton position measurement, 3D laser scanning measurement, and traditional vision schemes. However, these schemes always suffer from low adaptability and reliability. Recently, deep learning approaches have shown significant achievement for many tasks such as image detection, segmentation and natural language processing(NLP). CNN-based models on visual techniques show great application prospects in carton detection or segmentation tasks under various scenarios.

The training and evaluation of an excellent task-wise deep network are strongly tied to the construction of a large-scale dataset. Building CNN-based models to detect or segment cartons requires a large scale dataset of the cartons with scenario constraints. At present, several popular datasets are publicly available for CNN-based models development, \ie, MS COCO\cite{Lin2014Microsoft}, PASCAL VOC\cite{everingham2010pascal}, ImageNet\cite{deng2009imagenet} and Open Image\cite{2020The}. Nonetheless, most of them do not contain the category of carton. ImageNet and Open Image include related images but their number are small. Furthermore, images in these benchmarks are downloaded from the Internet and each image only cover several instances, making it difficult to simulate the various arrangements of cartons in a stacked scenario.

To fill the shortage of carton datasets, we establish a Stacked Carton Dataset(SCD) with a large scale of images only containing cartons and a wide variety of backgrounds. To further mimic different carton packing scenarios, the number of cartons in the image is arranged arbitrarily from sparse to dense distribution. we split the SCD into two subsets(called Live Stacked Carton Dataset(LSCD) and Online Stacked Carton Dataset(OSCD)) that are collected from the warehouse and downloaded from the Internet respectively. To ensure that cartons in images have rich texture information and application occasion, different logistics and storage locations are selected to collect image data, \ie, integrated e-commerce logistics warehouse, logistics transfer station, drug e-commerce logistics warehouse and fruit wholesale market, as well as the Internet to downloading images. The collecting spots of the online subset are also numerous to be greedily used for pre-training followed by fine-tuning a model for specific carton scenario and performing transfer learning, etc. As the most important part of the dataset, pixel-wise annotation is carried out for more precise localization. Particularly, the outer/inner and occlusion information are provided by dividing a carton into four categories including Carton-inner-all, Carton-inner-occlusion, Carton-outer-all and Carton-outer-occlusion. These bonus annotations have great application potential, for example, it is easy to determine which box can be grabbed by turning a logical problem into a classification problem.

SCD places unique challenges to computer vision research on detection\cite{2017Faster,Lin2017Focal,tian2019fcos} or instance segmentation\cite{fcn,he2017mask,yolact,polarmask,solo}. To evaluate the SCD, we compare several CNN-based detectors including single-stage\cite{Lin2017Focal,tian2019fcos} and two-stage frameworks\cite{2017Faster,he2017mask}. Then, we choose RetinaNet\cite{Lin2017Focal} as the baseline on our dataset because of its comprehensive performance and speed. However, RetinaNet suffers from a serious imbalance\cite{2020IoU,2018Acquisition} between classification and localization accuracy. In other words, the result of classification scores and localization quality is not equal absolutely and their relationship is even nonlinear. Nevertheless, RetinaNet coercively adopts classification results to mirror localization accuracy to participate in the NMS procedure, leading to sub-par performance. In this case, the predicting bounding boxes with higher localization accuracy may be filtered, while the inferior ones are yet reserved. To ease this situation, IoU-Net\cite{2018Acquisition} and IoU-aware RetinaNet\cite{2020IoU} are proposed to directly predict the IoU of samples. VarifocalNet\cite{2020VarifocalNet} and generalized focal loss\cite{2020Generalized} adopt an IoU-aware classification loss to predict localization confidence, but they give up the classification confidence which is essential in some specific occasions, such as automatic pickup of goods and tumor detection. To deal with this issue, we propose Offset Prediction between Classification and Localization module(OPCL) to bridge the gap from classification to localization quality while retaining classification head, and the gaps are adopted to modify classification confidence. 

When stacked cartons are detected, repetitive textures are often everywhere because cartons are often stacked together for the same purpose, which confuses the model to make wrong judgments. To eliminate the dilemma, we utilize the boundary information including the contour of the individual box and the topology of carton placement. Owing to the pixel-wise annotation, the boundary of each instance can be simply extracted without any other annotations. We propose a Boundary Guided Supervision module(BGS) to supervise and urge detectors to pay more attention to boundary information, which helps decouple repeated text and does not add computation overhead during inference.

The main contributions of the paper are summarized as follows:
 \begin{itemize}
  \setlength{\itemsep}{1pt}
  \setlength{\parskip}{0pt}
  \setlength{\parsep}{0pt}

 \item We establish a large-scale Stack Carton Dataset(SCD), which contains two subsets collected from the warehouse and downloaded from the Internet. Images collected on the site in different scenarios have rich texture information and high resolution while the part of the Internet can be used to pre-training and transfer learning.
 \item Pixel-wise annotation is applied to label instances in SCD. MS COCO\cite{Lin2014Microsoft} and PASCAL VOC\cite{everingham2010pascal} format annotation are also provided for detection and segmentation tasks. Four labels that contain inside/outside information and occlusion information of the carton are provided to cater to the operation requirement of carton goods. To best of our knowledge, SCD is the first public image dataset on the carton in diverse scenarios, which provides the premise for the carton cargo detection based on deep learning.
 \item We provide a baseline on SCD for evaluation by using several state-of-the-art deep learning models.
 \item We propose two novel module plugged in RetinaNet. Offset Prediction between Classification and Localization module(OPCL) is proposed to deal with the extreme imbalance of classification score and localization confidence with respect to samples. Boundary Guided Supervision module(BGS) is introduced to extract and utilize boundary information of cartons, which guides model to pay more attention to boundary information so as to promote carton localization.
\end{itemize}

The paper is organized as follows. Related works are discussed in Section~\ref{section:2}. The construction details and statistics of the SCD are introduced in Section~\ref{section:3}. The used carton detector and its novel modules are introduced in Section~\ref{section:4}. Extensive experimental results are published in Section~\ref{section:5}. Finally, conclusions are summarized in Section~\ref{section:6}.


\section{Related Work}
\label{section:2}

We first review several popular datasets and some of them contain carton images. Then several methods for the imbalance issues between classification and localization are discussed, and finally some related works for predicting target map are introduced.

\textbf{Large-Scale Datasets.} The great development of deep neural networks is inseparable from the establishment of large datasets which provides lots of annotation information for the related tasks of training models. The MS COCO\cite{Lin2014Microsoft} dataset is the most popular benchmark for evaluating novel models and poses a huge challenge to researchers. There are 80 object classes, over 1.5 million object instances and more than 200,000 images in the COCO dataset which supports classification, detection, and segmentation tasks. The PASCAL VOC\cite{everingham2010pascal} contains two versions(VOC2007 and VOC2012) which include 9,963 and 22,263 images respectively. 20 object classes are included in the two versions and the PASCAL VOC support detection and semantic segmentation. However, both of them do not include the object class of cartons. As the most frequently used dataset for pre-training, ImageNet\cite{deng2009imagenet} provides image-level labels for 5,247 classes and a large scale of images up to 3,200,000. This dataset has greatly prompted the development of computer vision in recent years, but only 1,388 images include cartons and cannot be used for carton detection in the dense carton packing scenes. The latest Open Image dataset\cite{2020The} contains 9,052,839 images and provides 7,186 object classes for the classification task and 600 classes for the detection task. It includes box classes which contains carton instances but it doesn't subdivide into paper boxes and suffers from scarcity, inferior quality.

\textbf{Misalignment of Classification and Localization Accuracy.} As the indispensable module, NMS plays a significant role for detectors to remove duplicated predicting bounding boxes and is used in most state-of-the-art models\cite{2017Faster,tian2019fcos,Lin2017Focal,Liu2016SSD}. Soft-NMS\cite{2017Improving} is proposed to adopt a decay strategy of confidence which is easy to implement and boost the recall and accuracy. Both NMS and Soft-NMS regard the classification confidence as the corresponding localization accuracy which does not give full play to the ability of regression and leads to suboptimal performance. To ease the misalignment, several works utilize an extra branch to predict various types of criterion which is integrated into classification score to simulate the localization quality and participate in the process of removing the repeated boxes. IoU-Net\cite{2018Acquisition} and IoU-Aware RetinaNet\cite{2020IoU} directly predict IoU between the predictive boxes and corresponding ground truth boxes in two-stage and single-stage models respectively. As an excellent anchor-free detector, FCOS\cite{tian2019fcos} predicts centerness scores to suppress the low-quality detections. Instead of adding an extra branch, IoU-balanced classification loss\cite{2019IoU} and PISA\cite{2019Prime} adopt respectively the normalized IoU and sorted score to reweight classification loss based on their localization accuracy, which strengthens the correlation of classification confidence and localization quality. Besides, several works aim to design a joint representation of localization accuracy and classification. VarifocalNet\cite{2020VarifocalNet} and generalized focal loss\cite{2020Generalized} modify the paradigm of focal loss and use IoU between bounding boxes and ground truth boxes instead of the target labels corresponding to the category of the positive samples. The IoU-aware methods naturally transfer classification scores to localization confidence but the pure classification confidence is discarded, which is fatal for tasks like diagnosing tumors.

\textbf{Map and Boundary Prediction.} To predict a target feature map, such as instance mask, boundary and texture, a popular method is to attach an extra head to predict the task-specific maps parallel to an excellent detector. YOLACT\cite{yolact}, as the first real-time approach method for instance segmentation, which is based on the RetinaNet\cite{Lin2017Focal}, attaches a head parallel to classification and localization tasks, generating a set of prototype masks and then predicting per-instance mask coefficients. Similarly, CondInst\cite{condinst} performs instance segmentation task by reformulating FCOS\cite{tian2019fcos} where a lightweight mask map is predicted from the lowest level of FPN while conditional convolution kernels are dynamically generated to integrate mask map and get vanilla instance-wise masks. PolarMask\cite{polarmask} uses polar representation to model the outer contour of instance and transforms per-pixel mask prediction problem to distance regression problem. Our BGS also attaches an extra head to predict the boundary of each instance, but our target is to guide detectors to pay more attention to contour information for more precisely localization. It's worth noting that we dismantle the prediction head during inference so that there is no extra computation.

\begin{figure}[t]
\begin{center}
\includegraphics[width=\linewidth]{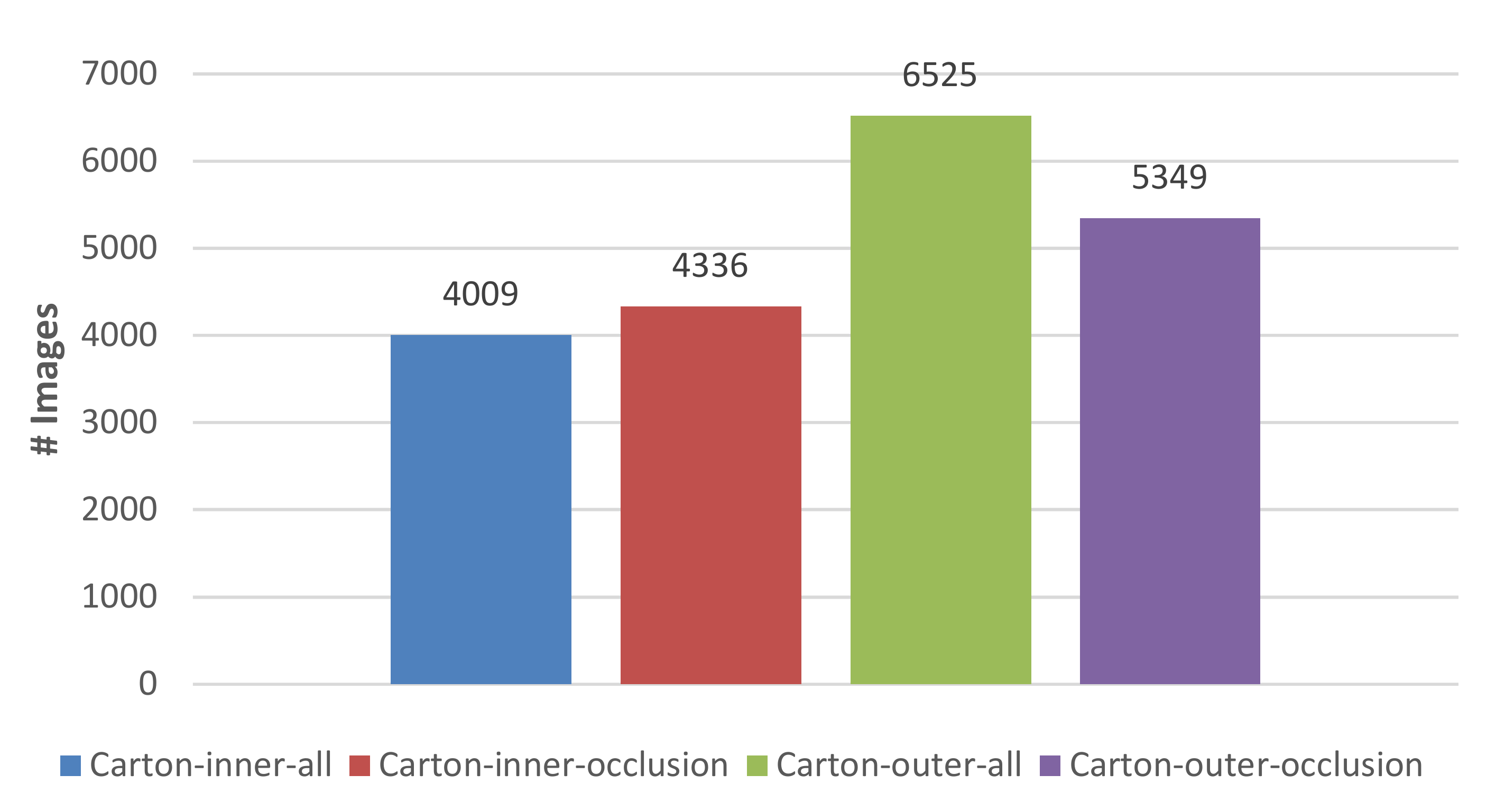}
\caption{The numbers of images with respective to 4 labels in the LSCD, which includes Carton-inner-all,  Carton-inner-occlusion, Carton-outer-all and Carton-outer-occlusion.}
\label{fourclassimages}
\end{center}
\end{figure}

\begin{figure}
\begin{center}
\includegraphics[width=\linewidth]{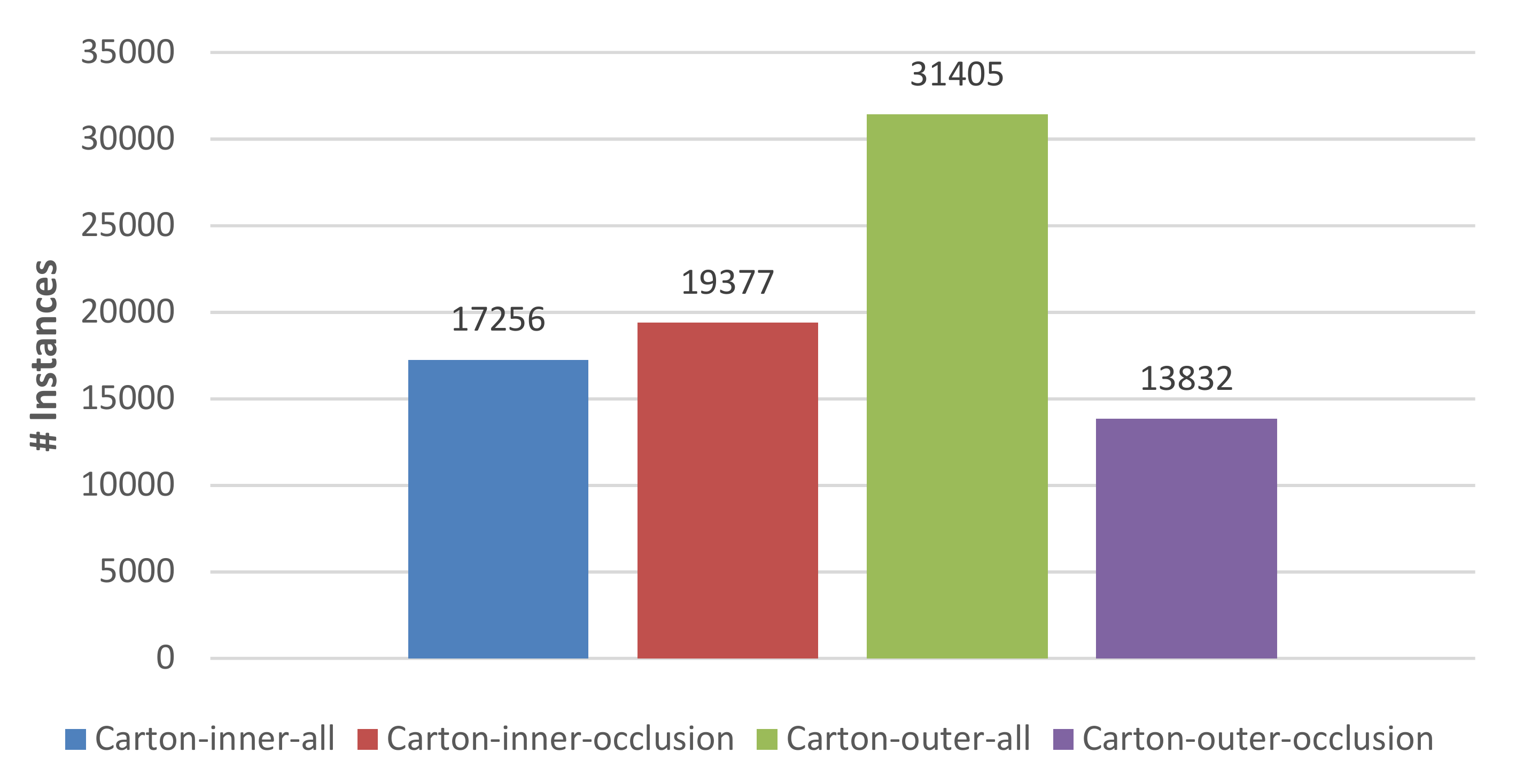}
\caption{The numbers of instances with respective to 4 labels in the LSCD, which includes Carton-inner-all,  Carton-inner-occlusion, Carton-outer-all and Carton-outer-occlusion.}
\label{fourclassinstances}
\end{center}
\end{figure}

\begin{figure*}
\begin{center}
	\frame{\includegraphics[width=1\linewidth]{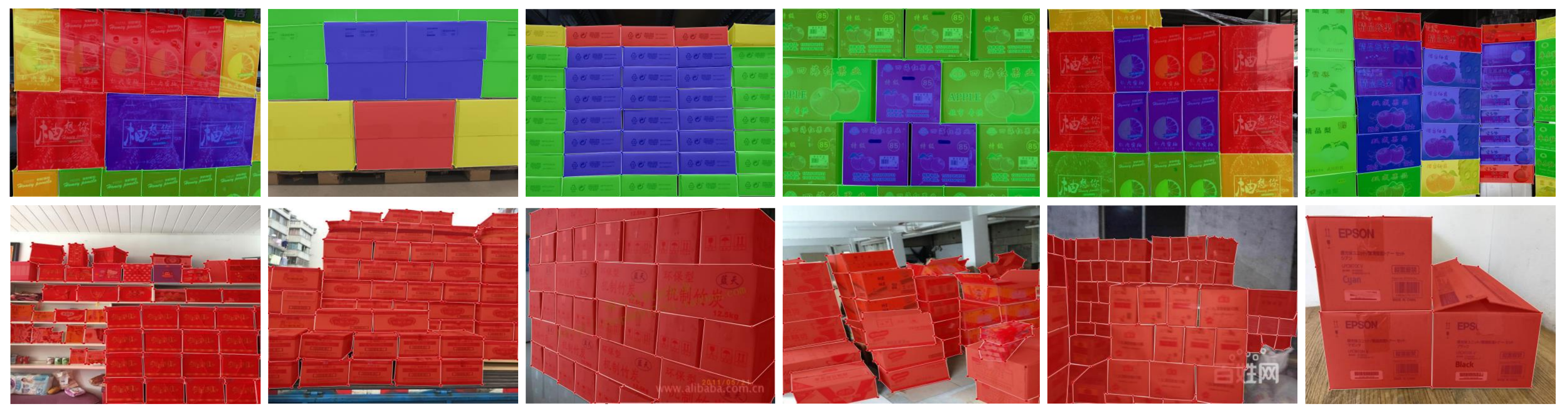}}
\end{center}
\vspace{-1em}
\caption{Example of instance annotation in SCD. The first line represents the style of four labels with respect to LSCD while the second line illustrates the style of one label in OSCD. In terms of first line, \textcolor{blue}{blue}, \textcolor{green}{green}, \textcolor{red}{red} and \textcolor{yellow}{yellow} represent \textcolor{blue}{Carton-inner-all}, \textcolor{green}{Carton-inner-occlusion}, \textcolor{red}{Carton-outer-all} and \textcolor{yellow}{Carton-outer-occlusion} respectively.}
\label{fig:annotation}
\end{figure*}

\section{Stacked Carton Dataset}
\label{section:3}

The popular datasets\cite{Lin2014Microsoft,everingham2010pascal,cordts2016cityscapes,deng2009imagenet,2020The} have been adopted to verify the performance of deep networks and have made great contributions to the development of computer vision. Regrettably, these benchmarks do not include carton scenes which are crucial for the application of logistics, transportation and robotics. In this section, we introduce a new large-scale Stacked Carton Dataset(SCD) focusing on detection and semantic understanding of carton, which is suitable for various carton distribution scenarios and has several appealing properties. First, we introduce the details of image collection and split, and then image annotation is described. Finally, the dataset statistics and properties are illustrated.

\subsection{Data Collection}
The source of data collection includes two aspects: online downloading and on-site shooting. In terms of images on the web, it is easy to find a large number images of carton because boxes are widely used in daily life, logistics, and other scenarios. Concretely, we integrated a batch of seed keywords and photos for matching and downloading, 8,401 images were collected after removing duplicate and filtering. However, data collected from the web has low resolution and even is mixed with noise such as watermarks and text. To mimic a real-world application scenario, we assembled 7735 images with high pixels and resolution from some typical carton stacking scenes, \ie, E-commerce warehouse, large wholesale market, integrated logistics warehousing and fruit market. There are several principles during data collection: 
\begin{itemize}
	\item The distance between the camera and stacked carton targets is controlled within 5 meters.
	\item The layout of different cartons in one image appears within the same distance scale. Try not to let cartons far away and extremely close appear on the same horizon.
	\item One photo has only one group of cartons. If several piles of cartons appear, the distance scale between the target and the camera is strictly limited.
	\item Try to shoot from a perspective perpendicular to a group of cartons.
\end{itemize}

Compared with ImageNet\cite{deng2009imagenet}, the source of our data is wider, the number of the image is several times more than it in terms of carton class and the number of instances in one photo is distributed from sparse to dense while one photo in the ImageNet benchmark only contains one or few cartons. Moreover, we give more meaningful labels for an instance to cater to other tasks described in section~\ref{section:3.3}.

	\begin{figure*}[!htb]\footnotesize
		\setlength{\abovecaptionskip}{0.cm}
		\setlength{\belowcaptionskip}{-0.cm}
	\centering
	\setlength{\tabcolsep}{1pt}
	\begin{tabular}{cccccccc}
		\includegraphics[width=0.2\textwidth,height=3cm]{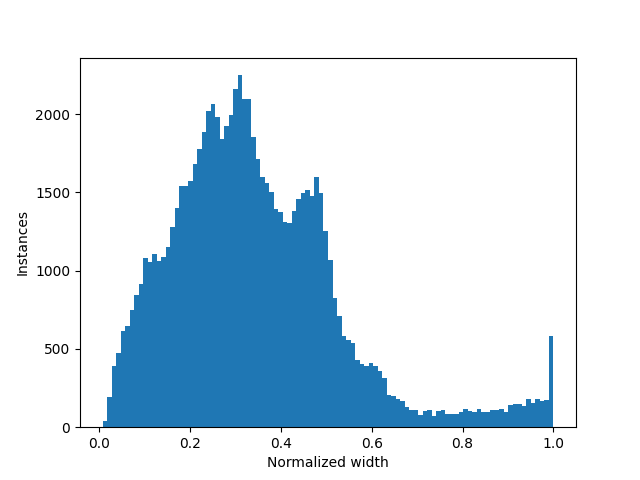}&
		\includegraphics[width=0.2\textwidth,height=3cm]{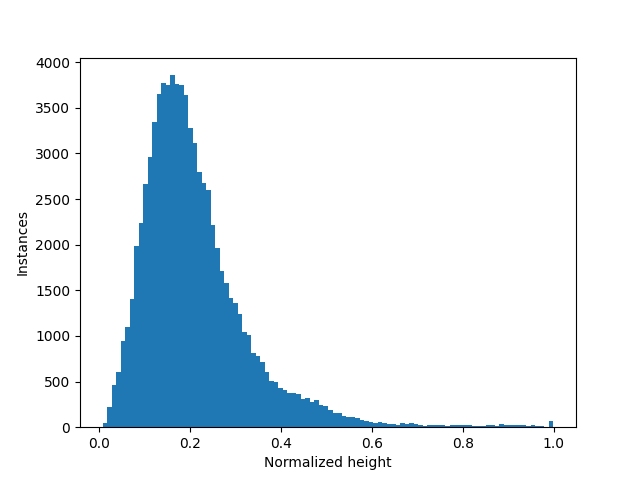}&
		\includegraphics[width=0.2\textwidth,height=3cm]{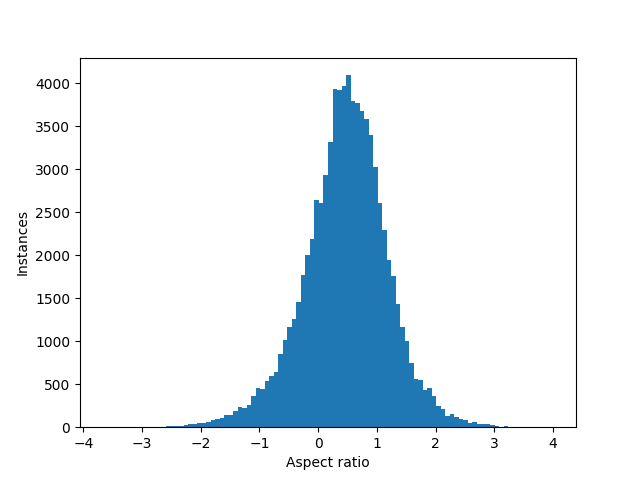}&
		\includegraphics[width=0.2\textwidth,height=3cm]{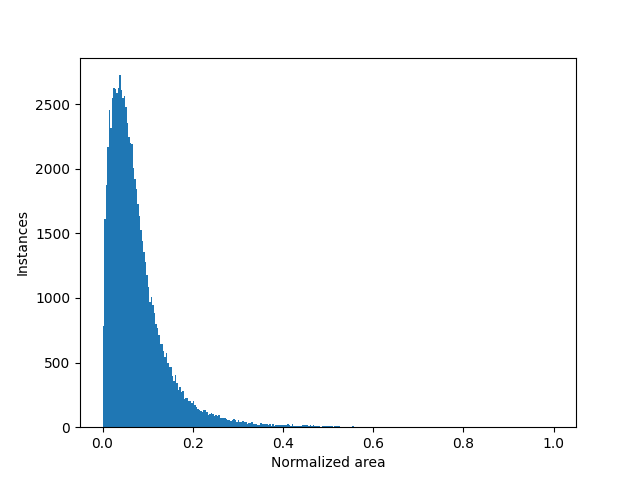}&
		\includegraphics[width=0.2\textwidth,height=3cm]{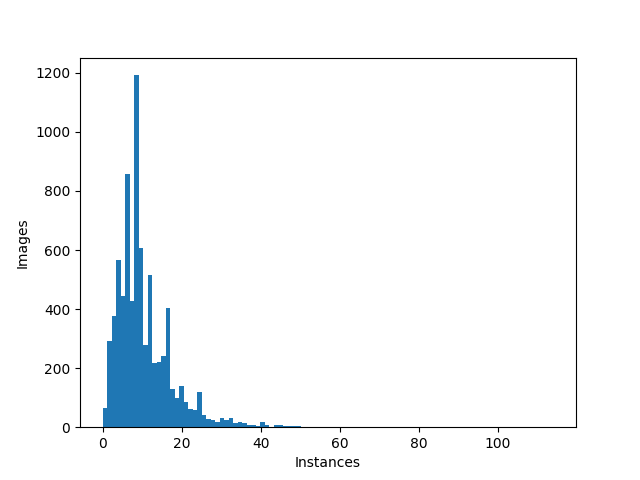}\\
		$(a)$ & $(b)$ & $(c)$ & $(d)$ & $(e)$ \\
		\includegraphics[width=0.2\textwidth,height=3cm]{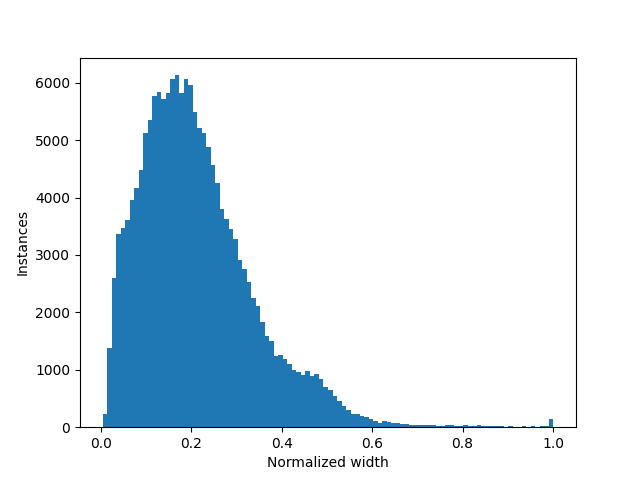}&
		\includegraphics[width=0.2\textwidth,height=3cm]{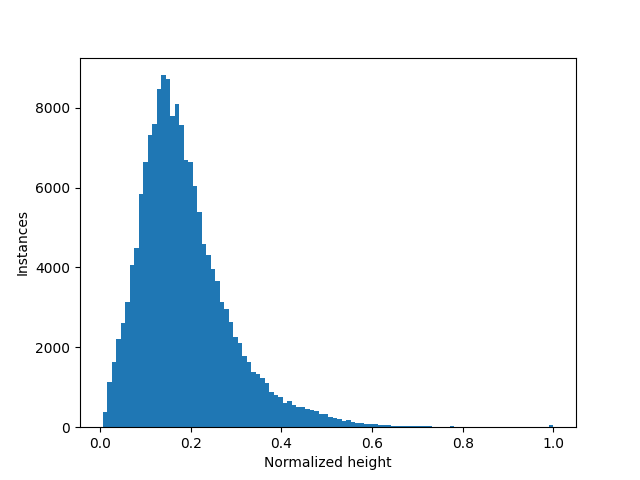}&
		\includegraphics[width=0.2\textwidth,height=3cm]{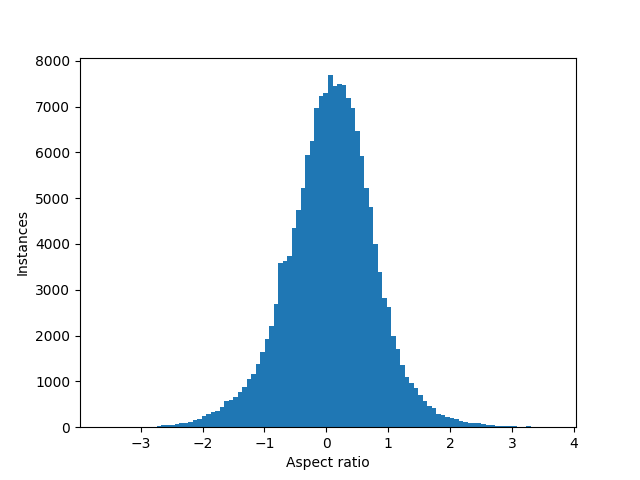}&
		\includegraphics[width=0.2\textwidth,height=3cm]{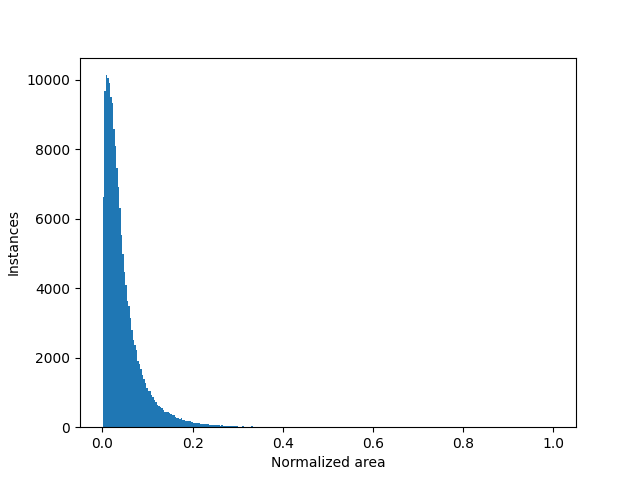}&
		\includegraphics[width=0.2\textwidth,height=3cm]{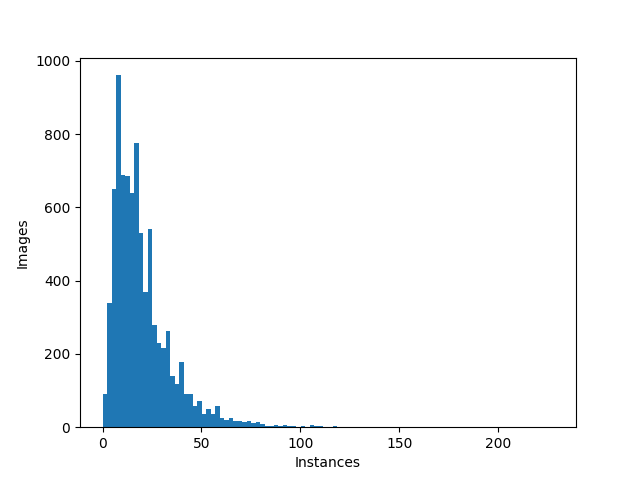}\\
		$(f)$ & $(g)$ & $(i)$ & $(z)$ & $(k)$ \\
	\end{tabular}
	\caption{The first line represents the statistical distribution of LSCD while the second line represents the statistical distribution of OSCD. The chart calculates the width, height, aspect ratio, pixel area and the number of objects in each image from left to right. Noting that the width, height and area of instance are all normalized by the width and height of corresponding image. Log function is adopted to normalize aspect ratio.}
	\label{fig:4}
\end{figure*}

\subsection{Data Splits}
To adapt to different tasks of diverse scenarios, we split the images into two subsets(called Live Stacked Carton Dataset(LSCD) and Online Stacked Carton Dataset(OSCD)) which all contain training and testing sets:
\begin{itemize}
	\item LSCD: It contains all images taken at the carton stacking loading and unloading scene. The images are randomly split into two parts including 6,735 images for training and 1,000 images for testing.
	\item OSCD: It contains all images collected from the Internet to provide a substantial number of images for pre-training or transfer learning tasks. Likewise, 1000 images are randomly selected for testing while the remaining 7401 images for training.
\end{itemize}

Although only two subsets are specified, there are several different combinations. For example, LSCD and OSCD are integrated for pre-training or the semantic knowledge achieving from OSCD is transfer to related carton tasks.

\subsection{Image Annotation}
\label{section:3.3}

The same as MS COCO\cite{Lin2014Microsoft}, we use the polygon box to mark object instances. Figure~\ref{fig:annotation} shows the pixel-wise annotations applied in LSCD(first line) and OSCD(second line) respectively. We define only one annotation label(Carton) for annotation for images collected from the Internet while four labels(Carton-inner-all, Carton-inner-occlusion, Carton-outer-all, and Carton-outer-occlusion) for images in LSCD. One label(Carton) is also supported in LSCD so that it is easy to combine with OSCD for training and validation. We use LabelMe project\cite{labelme} to implement these annotations and transfer this labeling information into MS COCO \cite{Lin2014Microsoft} and PASCAL VOC\cite{everingham2010pascal} paradigms.

These are several rules and corresponding properties for the design of these 4 labels:
\begin{itemize}
	\item \textbf{How to distinguish between inner and outer.} For a given carton in pile of cartons, the "inner" label is marked if the contour of instance is all contacted by cartons. The instance is truncated by image edges is also judged to be connected to the cartons. All cartons except the "inner" label are "outer". As shown in the first line of Figure~\ref{fig:annotation}, blue and green masks represent the "inner" while red and yellow ones belong to the "outer".
	\item \textbf{How to distinguish between all and occlusion.} Given a carton, as long as there is a complete face, the instance belongs to "all" label while instance has no complete face is assigned "occlusion" label. Similarly, when a surface is truncated by image edges which is a deficient face. As the first line of figure~\ref{fig:annotation} shows, blue and red masks represent "all" while green and yellow ones are belong to "occlusion".
\end{itemize}


\subsection{Dataset Statistics}
In Table~\ref{tab:1}, our dataset is compared with existing popular datasets containing related carton images. It demonstrates that ImageNet\cite{deng2009imagenet} and Open Image\cite{2020The} only have a small number of carton images. In the Open Image dataset, only the label of "box" is provided which contains part of the sub-data of the carton. Therefore, it can not be utilized for related carton tasks. Our carton dataset SCD includes a total number of 16,136 images collected from site and Internet which makes the images in SCD more diverse. Furthermore, topology information for the placement of the carton is provided in labels, which is important for some logistics scenarios. Finally, SCD is suitable for both detection task and segmentation task. By comparison, SCD has the advantages of a substantial quantity, rich annotated information, and superior quality. Thus, SCD is suitable for diverse tasks which is related to carton detection and segmentation.

Figure~\ref{fourclassinstances} and Figure~\ref{fourclassimages} describe the relationship among the number of images, instances, and the 4 types of labels in terms of LSCD. The number of instances and images with respect to the "Carton-outer-all" label is more than that of the other labels. Figure~\ref{fig:4} illustrates the distribution of instances in all images with respect to LSCD and OSCD, which contains the width, height, aspect ratio, and pixel area of a marked rectangular box. Width and height are normalized by the width and height of the corresponding images respectively. The Log function is adopted to normalize aspect ratio because it is boundless. The first 4 histograms of the two rows indicate that width, height, and area distribution is close to the gaussian model and most instances have medium size. In Figure~\ref{fig:4} (a), the reason why the spike appears near the horizontal coordinate 1 is that a few slender cartons on LSCD are specially collected. Finally, (e) and (f) represent the number of cartons in each image, which demonstrates that the dataset obeys from sparse to dense distribution, and the maximum number of instances in one image is over 100, which is a challenge for instance detection and segmentation.

\subsection{Property}
\begin{figure*}[!bh]
\begin{center}
	\frame{\includegraphics[width=1\linewidth]{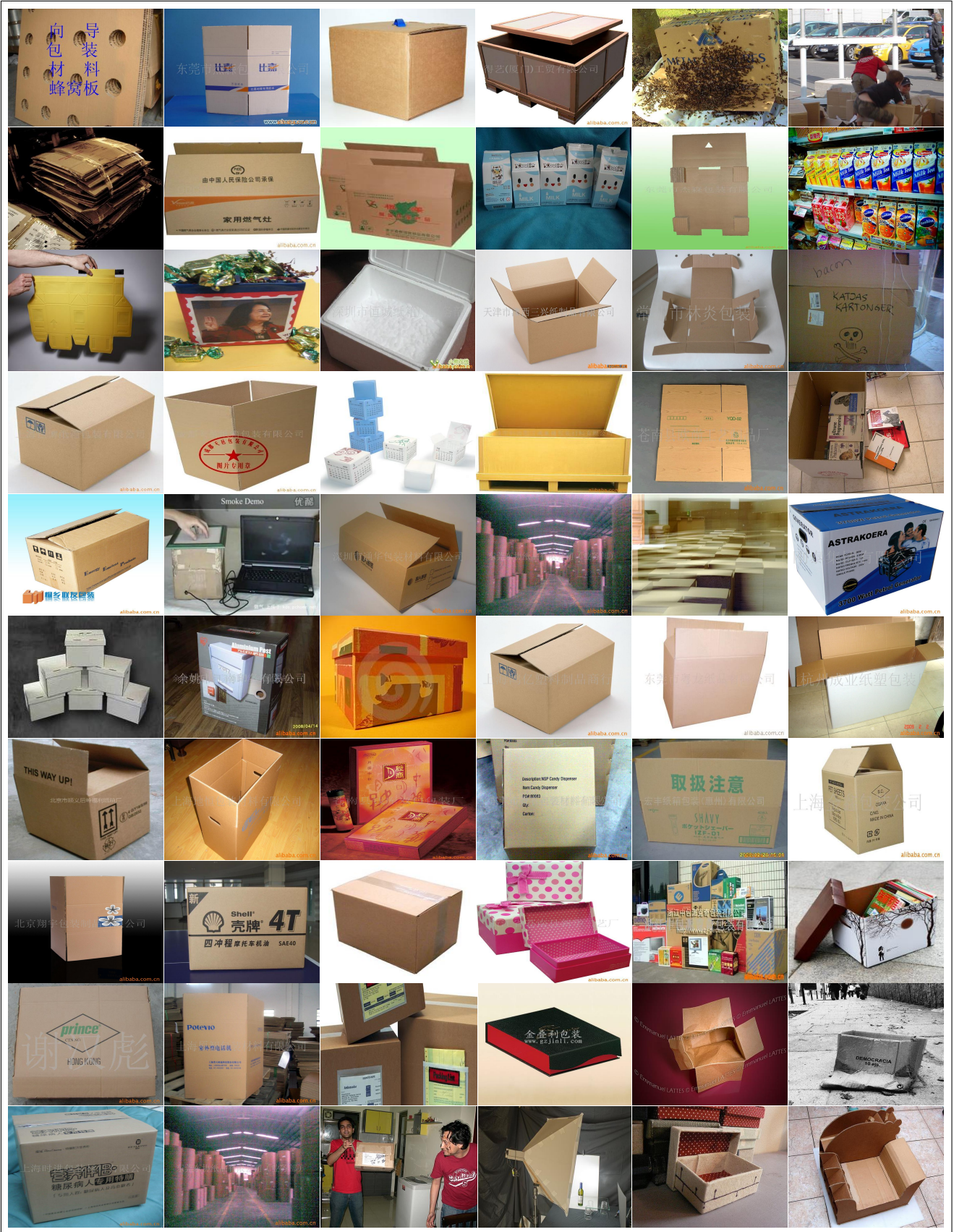}}
\end{center}
\vspace{-1em}
\caption{Randomly sampled images related to cartons on ImageNet dataset.}
\label{fig:imagenet_carton}
\end{figure*}

\begin{figure*}[!bh]
\begin{center}
	\frame{\includegraphics[width=1\linewidth]{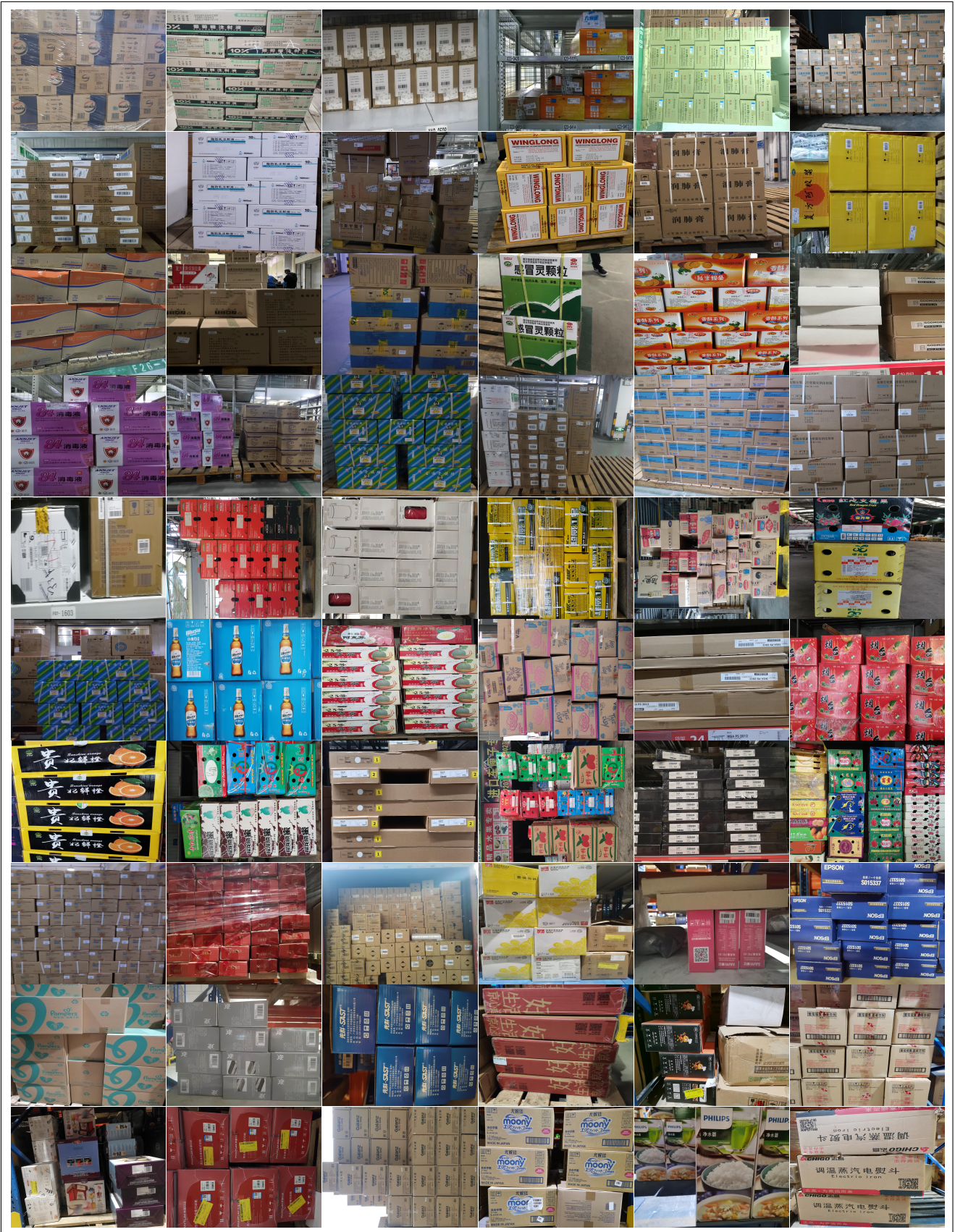}}
\end{center}
\vspace{-1em}
\caption{Randomly sampled images related to cartons on SCD.}
\label{fig:scd_carton}
\end{figure*}

As shown in Figure~\ref{fig:imagenet_carton} and Figure~\ref{fig:scd_carton} at the end, some raw images of ImageNet with respect to carton and our SCD are shown by random sampling. By comparison, it highlights the properties of LSCD which are suitable for logistics scene. We list some corresponding property between ImageNet and LSCD(For simplicity, A represents the properties of ImageNet while B is responsible for LSCD):
\begin{itemize}
	\item \textbf{Distribution of instances}. A: instances almost distribute in the center of image. B: The cartons are distributed in arbitrary positions of the photo.
	\item \textbf{Resolution}. A: blurred and low resolution. B: clear and high resolution.
	\item \textbf{Noise}. A: all images are downloaded from the Internet so that some noises, such as watermark are attached. B: all Images are collected at the scene and almost no noise so that it can mimic real scene well.
	\item \textbf{Texture}. A: mostly brown cartons and white background. B: both the cartons and the background have rich textures.
	\item \textbf{purity}. A: it mixes with carton-like box, such as wood, foam, and iron boxes and some carton are squashed and wrinkled. B: it only contains cartons which are unfolded into boxes.
\end{itemize}


\section{Methodology}
\label{section:4}

Comparing with several CNN-based models, RetinaNet\cite{Lin2017Focal} is selected for detection because of its better trade-off between speed and performance. In this section, we first review the RetinaNet and then introduce the novel OPCL(Offset Prediction between Classification and Localization), which is a unified and lightweight head attached to RetinaNet. Finally, BGS(Boundary Supervision Supervision) is described to further boost the performance of the model.

\begin{figure*}[!tbh]
\begin{center}
	\frame{\includegraphics[width=1\linewidth]{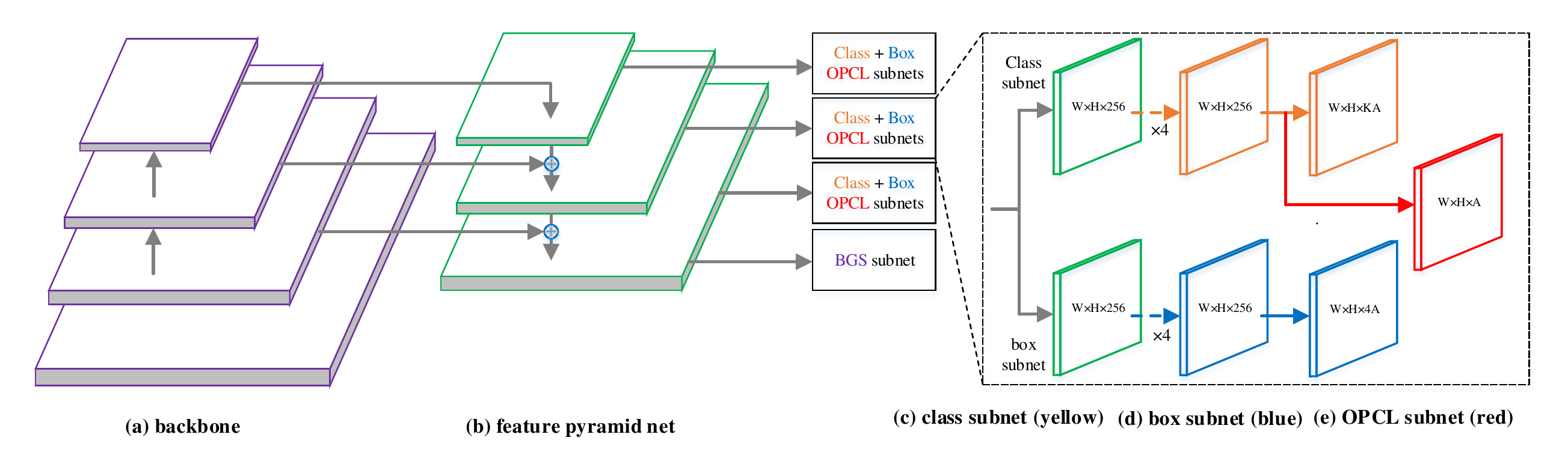}}
\end{center}
\vspace{-1em}
\caption{Illustration of our detector which contain backbone, feature pyramid net(FPN) and prediction head. Prediction head include classification, regression and Offsets Prediction between Classification and localization(OPCL) head. Boundary Guided Supervision module(BGS) is only attached at ${{P}_{3}}$ of FPN.}
\label{fig:model}
\end{figure*}

\subsection{RetinaNet}

RetinaNet is a representative single-shot detector, as shown in Figure~\ref{fig:model}, whose network consists of a backbone and a feature pyramid network(FPN) with five detection heads. FPN is constructed by taking the backbone network with levels from ${C}_{3}$ to ${C}_{5}$(${C}_{i}$ is the output of corresponding ResNet residual stage) as input and fusing features map with a top-down pathway and lateral connection. Each level of the pyramid is used for detecting objects at different scales respectively. The multi-scale feature maps are tiled with densely hand-made anchors, which gives the detector the ability to predict various size instances. Its classification and localization subnets are both small fully convolutional networks consisting of 4 stacked convolution layers attached to ${P}_{l}$(${P}_{l}$ represents the output of corresponding $\ell$-th level of FPN). The classification head predicts the probability of objects with respect to each anchor for each of the $K$ object classes while the localization head predicts the 4-dimensional class-agnostic offsets between the anchor and the ground-truth box. Owing to the imbalance between hard and easy examples and the imbalance between foreground and background class, focal loss is adopted to ease this problem. The localization task uses Smooth ${\ell}_{1}$ function to regress the 4 offsets of anchors.

\subsection{Offset Prediction between Classification and Localization}

\textbf{Learning to predict offset.} To bridge the gap from classification score to localization confidence, we adopt IoU between predicted bounding boxes and corresponding ground-truth boxes as localization confidence as the same as \cite{2019IoU,2018Acquisition}. Fortunately, the criterion(IoU) is easily calculated by the 4 predicted offsets between predicted boxes and corresponding anchors. After defining the target localization confidence, the question can be shifted to how to predict the gap directly so that the target gap is naturally achieved. Next, it is important to choose which head to attach the OPCL. We select to construct the gap prediction layer parallel with classification layer. Actually, the regression head can also be attached, but we find that it gets a suboptimal performance. The style of concatenation between the parallel heads are also implemented but the performance is the same as the first one. Therefore, we choose the first style because of the negligible computation and better performance. As shown in Figure~\ref{fig:model}, a 3×3 conv layer with A filters is applied to output the gap prediction per spatial location(A represents the number of anchors on each location). 

\textbf{Training and inference.} The loss functions for classification and localization are the same as RetinaNet. Focal loss and smooth ${\ell}_{1}$ loss are applied to these two branches respectively. As for the gap prediction branch, the binary cross-entropy loss(BCE) is adopted, whose performance is better than other loss functions, \ie, L1 Loss or L2 loss. Equ.\ref{eq:gap} and Equ.\ref{eq:bce} show the loss functions of gap prediction. For simplicity, ${C}_{loc}$ represents the i-th positive anchor in Equ.\ref{eq:bce}. During training, we add predicted gaps with corresponding classification scores and then induce sigmoid activation. The correctional criterion simulates the localization confidences and then is fed into BCE loss together with corresponding IoU. Similar to regression loss, only positive samples are applied to predict the gap. It is worth noting that whether to calculate the gradient of ${{{L}}_{gap}}$ with respective to IoU get completely different results. By transferring the gradient of IoU backward, the localization accuracy is further improved. During inference, the predicted gaps are adopted to compensate for classification scores. Equ.\ref{eq:infe} shows that gaps are added to the classification score and then feed into the sigmoid function, which is actually the predicted localization confidence. However, we do not use synthesized confidence as a criterion to apply NMS procedure, instead, we define a controlling factor $\alpha$ to integrate the original classification score and corresponding predicted localization confidence as ${S}_{det}$ which is the vanilla criterion in NMS procedure.

\begin{equation}
    \label{eq:gap}
    {{C}_{loc}}={{C}_{cls}}+{{C}_{gap}}
\end{equation}

\begin{equation}
    \label{eq:bce}
    {{L}_{gap}}=\frac{1}{{{N}_{pos}}}\sum\limits_{i\in {{N}_{pos}}}^{{}}{BCE\left(Sigmoid{\left( {{C}_{loc}}\right)},{{IoU}_{i}} \right)}
\end{equation}

\begin{equation}
    \label{eq:infe}
    {{S}_{\det }}=Sigmoid{{\left( {{C}_{cls}}+{{C}_{gap}} \right)}^{\alpha }}\bullet {{C}_{cls}}^{\left( 1-\alpha  \right)}
\end{equation}

\begin{figure}
\begin{center}
\includegraphics[width=\linewidth]{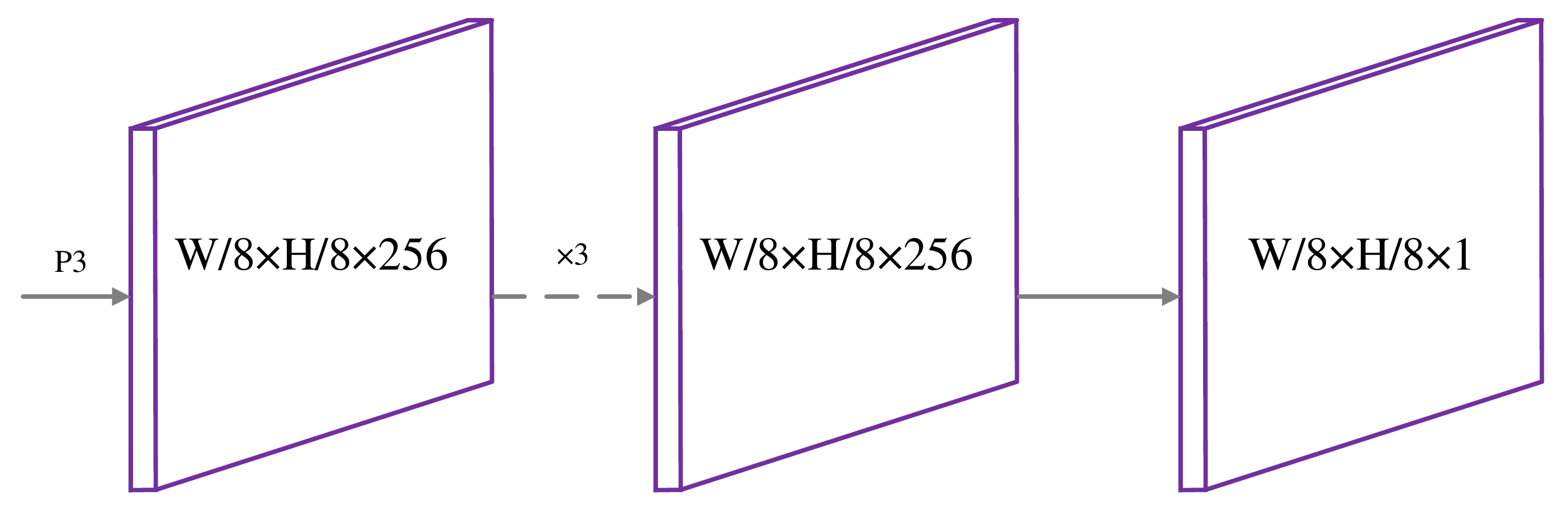}
\caption{Illustration the architecture of BGS head. Arrows
indicate 3 × 3 conv layer while the final map represents predicted boundary map whose size is $W/8 * H/8$}
\label{bgs}
\end{center}
\end{figure}

\begin{figure}
\begin{center}
\includegraphics[width=\linewidth]{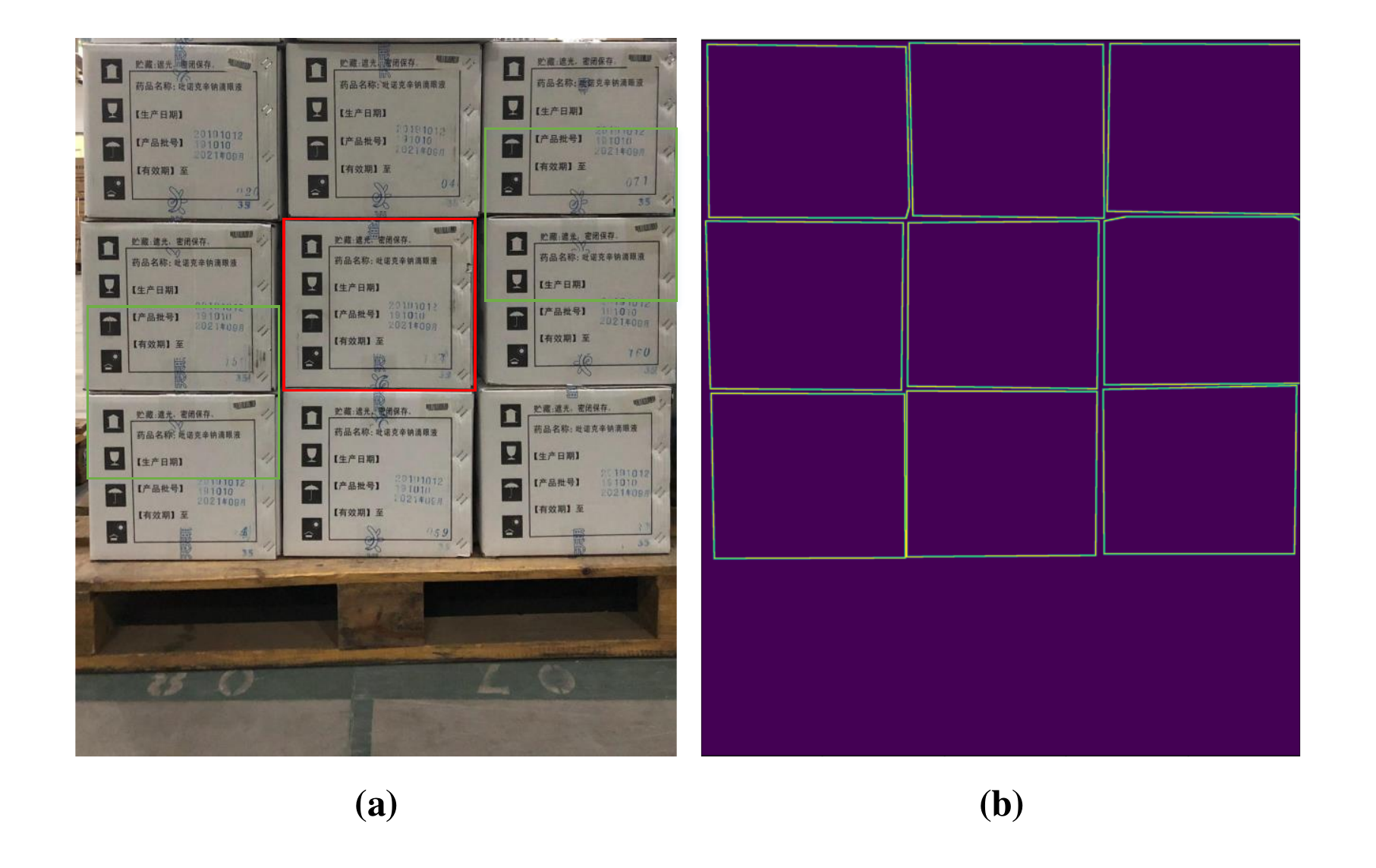}
\caption{Illustration of the target boundary map which is extracted naturally from pixel-wise annotation. red and green boxes in (a) represent correctly positioned boxes and error boxes that are confused by repeated texture. (b) is the binary image where the pixel value of the contour is 1.}
\label{boundary}
\end{center}
\end{figure}

\subsection{Boundary Guided Supervision}
\label{section:4.3}

\textbf{Architecture.} Boundary Guided Supervision(BGS) is a simple, unified network for guiding detector to pay more attention to boundary clue. As shown in (a) and (b) of Figure~\ref{boundary}, stacked cartons often have repetitive texture information which confuses the model to make wrong positioning decisions, such as the emergence of the middle box(the green boxes in Figure~\ref{boundary} (a)). Supervising model to generate target contour is conducive to decouple the high density and stack structure between cartons so as to further promote instances localization. We implement BGS as an FCN whose last layer has a channel. As illustrated in Figure~\ref{fig:4}, ${{P}_{3}}$ is the lowest level of Feature Pyramid Network(FPN) and is selected to construct the branch because of its the high resolution and detailed boundary information. The architecture is shown in figure~\ref{bgs}. The same as two head in RetinaNet\cite{Lin2017Focal}, 4 stacked 3×3 conv layer with C channels followed by ReLU are applied to ${{P}_{3}}$. Finally, 3×3 conv layer and sigmoid activations are attached in turn to output the prediction boundary map with 1/8 resolution of the input image.

\textbf{Loss function.} There are many choices for the loss of BGS, such as Binary Cross Entropy(BCE), Focal Loss(FL)\cite{Lin2017Focal} or Dice Loss(DL)\cite{dice}. How to choose a better loss depends on its effect on the performance improvements from supervision. Due to the natural imbalance between boundary and background pixels in prediction scenarios, where the pixels of boundaries always occupy few parts(See Figure~\ref{boundary} (b)), FL and DL may be more suitable for the implementation of BGS. The upper bound of FL is verified to be more advantageous than that of the other two losses mentioned above, as shown in Figure~\ref{bgs_loss}. Therefore, the vanilla loss for BGS is:

\begin{equation}
    \label{eq:infe}
    {{L}_{bgs}}=\frac{1}{{{N}_{pos}}}\sum\limits_{i\in N}^{{}}{\left( -{{\alpha }_{t}}{{\left( 1-{{p}_{t}} \right)}^{\gamma }}\log \left( {{p}_{t}} \right) \right)}
\end{equation}

\begin{equation}
    \label{eq:infe}
    {{p}_{t}}=\left\{ \begin{matrix}
   \begin{matrix}
   p & ,\text{   boundary pixel}  \\
\end{matrix}  \\
   \begin{matrix}
   1-p & ,\text{ background pixel}  \\
\end{matrix}  \\
\end{matrix} \right.
\end{equation}

\begin{equation}
    \label{eq:infe}
    {{\alpha}_{t}}=\left\{ \begin{matrix}
   \begin{matrix}
   \alpha & ,\text{   boundary pixel}  \\
\end{matrix}  \\
   \begin{matrix}
   1-\alpha & ,\text{ background pixel}  \\
\end{matrix}  \\
\end{matrix} \right.
\end{equation}
The loss of ${{L}_{bgs}}$ calculates both positive and negative samples but only positive samples are used to its weight. p is the prediction pixel value of boundary map. ${\alpha}_{t}$ is used to balance positive and negative pixels. Noting that the network of BGS is removed during inference so that there is no extra computation overhead.


\section{Experimental Results}
\label{section:5}

In this section, Several state-of-the-art detectors such as RetinaNet\cite{Lin2017Focal}, Faster R-CNN\cite{2017Faster} and FCOS\cite{tian2019fcos} are trained on SCD as baselines. Then our proposed modules(OPCL and BGS) are attached to the RetinaNet to demonstrate their superiority. To show the generalization ability of OPCL, several tables are stated while two popular datasets(MS COCO and PASCAL VOC) are evaluated too.
\\\\

\subsection{Experiment Setting}

\textbf{Dataset:} We evaluate the performance of our vanilla model and design ablation studies on our dataset SCD and other challenging datasets including PASCAL VOC and MS COCO. 

\emph{SCD}: The dataset is split into LSCD and OSCD which are formed by live photos and Internet photos respectively. LSCD contains 6636 training and 1000 testing images while OSCD includes 7401 and 1000 images for training and testing respectively. In addition, LSCD supports two different experiments because of its two types of labels.

\emph{PASCAL VOC 2007/2012:} The Pascal Visual Object Classes
(VOC)~\cite{Everingham2010The} benchmark is one of the most widely used
datasets for classification, object detection, and semantic segmentation. PASCAL VOC 2007 consists of 5011 images for training (\textit{VOC2007 trainval}) and 4952 images for testing(\textit{VOC2007 test}). 
PASCAL VOC 2012 consists of 11540 images for training (\textit{VOC2012 trainval}) and 10591 images for testing(\textit{VOC2012 test}). There are totally 20 categories of objects and all these objects have been annotated with bounding boxes.

\emph{MS COCO:} Recently, almost all the detection models take Microsoft Common Objects in Context (MS-COCO)~\cite{Lin2014Microsoft} for image captioning, recognition, detection and segmentation testing. It consists of 118k images for training (\textit{train-2017}), 5k images for validation (\textit{val-2017}), 20k images for testing (\textit{test-dev}) and totally over 500k annotated object instances from 80 categories.

\textbf{Evaluation Protocol.} In this paper, we adopt the same performance measurement as the MS COCO Challenge~\cite{deng2009imagenet} to report our results. This includes the calculation of mean Average Precision (mAP) over different class labels for a specific value of $IoU$ threshold in order to determine true positives and false positives. The main performance measurement used in this benchmark is shown by \textbf{AP}, which is averaged mAP across different value of $IoU$ thresholds, \ie $IoU = \{.5, .55, \cdots, .95\}$. Then, $\text{A}{{\text{P}}_{S}}$ (AP for small scales), $\text{A}{{\text{P}}_{M}}$ (AP for medium scales) and $\text{A}{{\text{P}}_{L}}$ (AP for large scales) are evaluated.
                        
\textbf{Implementation Details.} All experiments are implemented based on PyTorch and MMDetection \cite{chen2019mmdetection}. ResNet-50, ResNet-101 and ResNeXt-101 are used as backbones in RetinaNet\cite{Lin2017Focal}, which are pre-trained in ImageNet. A mini-batch of 4 images per GPU is used during training RetinaNet and other detectors, thus making a total mini-batch of 8 images on 2 GPUs. The synchronized Stochastic Gradient Descent (SGD) is used for network optimization. The weight decay of 0.0001 and the momentum of 0.9 are adopted. A linear scaling rule is carried out to set the learning rate during training (0.005 in RetinaNet) and a linear warm up strategy is adopted in the first 500 iterations. Except that the learning rate changes linearly with mini-Batch, all parameter settings are consistent with the default settings of MMDetection.

\subsection{Baselines and Comparisons}
\textbf{Baselines on Stacked Carton Dataset.} We report the results and comparisons of three state-of-the-art methods including RetinaNet, FCOS and Faster R-CNN on SCD. As shown in Table~\ref{table:baselines}, RetinaNet with GIoU loss(5 is given for the weight of localization loss) performs best in all subsets of SCD with one or four labels. By comparison, RetinaNet is a detector with the best trade-off between performance and speed. Thus RetinaNet is selected as the baseline to evaluate our novel modules and implement ablation studies. By pre-training all detectors in OSCD and then fine-tuning in LSCD with one or four labels, the performance of detectors can be significantly improved by $1.4\%\sim7.3\%$ AP, which indicates that the dataset collects from the Internet is effective to provide transcendental knowledge for carton detection in different scenarios.

\begin{table}[!tb]\footnotesize
\centering
   \caption{\footnotesize Comparison of detection performance between three state-of-the-art methods on SCD. For the evaluation of LSCD, 1 and 4 labels are all evaluated. LSCD+OSCD means detector are firstly pre-trained in OSCD and then finetuned in LSCD. RetinaNet+ represents GIoU loss is used.}
 \begin{tabular}{c c c c c c} 
 \hline 
Dataset & labels & Model & $\text{mAP}$ & $\text{AP}_{50}$ & $\text{AP}_{75}$\\
\hline
OSCD & 1 & RetinaNet & 72.1 & 90.8 & 80.5 \\
OSCD & 1 & RetinaNet+ & 76.6 & 91.8 & 83.6\\
OSCD & 1 & FCOS & 72.8 & 91.1 & 80.6 \\
OSCD & 1 & Faster R-CNN & 69.0 & 90.1 & 77.8 \\
 \hline
LSCD & 1 & RetinaNet & 79.8 & 95.2 & 87.9 \\
LSCD & 1 & RetinaNet+ & 84.7 & 95.8 & 89.8 \\
LSCD & 1 & FCOS & 76.5  & 93.7 & 84.3 \\
LSCD & 1 & Faster R-CNN & 77.5 & 94.5 & 86.3 \\
 \hline
LSCD & 4 & RetinaNet & 65.7 & 80.4 & 73.0 \\
LSCD & 4 & RetinaNet+ & 69.9 & 80.0 &  74.9\\
LSCD & 4 & FCOS & 68.1 & 81.2 & 74.8 \\
LSCD & 4 & Faster R-CNN & 61.2 & 79.5 & 70.1 \\
 \hline
LSCD+OSCD & 1 & RetinaNet & 82.2 & 95.9 & 89.8 \\
LSCD+OSCD & 1 & RetinaNet+ & 86.1 & 96.3 & 91.2 \\
LSCD+OSCD & 1 & FCOS & 83.8 & 96.2 & 90.4 \\
LSCD+OSCD & 1 & Faster R-CNN & 80.6 & 95.7 & 89.2 \\
 \hline
LSCD+OSCD & 4 & RetinaNet &  67.4 & 80.8 & 74.1\\
LSCD+OSCD & 4 & RetinaNet+ & 71.5 & 80.9 & 76.4 \\
LSCD+OSCD & 4 & FCOS & 71.1 & 82.0 & 76.8 \\
LSCD+OSCD & 4 & Faster R-CNN & 64.7 & 81.2 & 73.7 \\
 \hline
 \end{tabular}
  \label{table:baselines}
\end{table}

\subsection{Ablation Studies of OPCL on SCD}

\textbf{The effectiveness factor of OPCL on different backbone.} Table~\ref{table:carton} shows the performance of the Offsets Prediction between Classification and Localization module(OPCL) which is embedded into RetinaNet on the LSCD. OPCL with different backbones has a considerable increase of 3.1\% - 4.7\% over the baseline in mAP. It is worth noting that whether to apply backpropagation on the gradient of target IoU can makes a significant performance difference. By reverse-passing the IoU gradient, it further improves the performance of OPCL by $1.8-3.3\%$.

\textbf{Which head to attach OPCL.} To investigate which head is suitable to attach OPCL, we construct it followed by classification head, regression head, and the combine head of the former two respectively. The first two styles use the final feature map of the 4 stacked maps before the prediction results. For the style of combine one, the feature maps of the former two styles are concatenated and then apply a 3×3 conv layer to get final offsets prediction. As shown in Table~\ref{table:layout}, OPCL which is attached on the classification head gets a better result and costs fewer parameters than the combined style.

\begin{table}[!tb]\footnotesize
\centering
   \caption{\footnotesize Comparison between three type of head to attach OPCL.}
 \begin{tabular}{c|cccccc} 
 \hline
Head & mAP & $\text{AP}_{50}$ &$\text{AP}_{60}$ & $\text{AP}_{70}$ &$\text{AP}_{80}$ &$\text{AP}_{90}$\\
 \hline
cls & 84.5 & 95.3 & 94.0 & 91.7 & 87.1 & 74.5\\
reg & 84.1 & 95.4 & 94.2 & 91.6 & 96.6 & 73.9\\
cls\&reg & 84.4 & 95.4 & 94.0 & 91.6 & 87.3 & 74.5\\
 \hline
 \end{tabular}
  \label{table:layout}
\end{table}

\begin{table*}[t]
\begin{center}
\centering
   \caption{\footnotesize Experimental results on LSCD. All the models are trained on LSCD trainval and evaluated on LSCD testing with the image scale of [600, 1000]. All the other settings are adopted as the same as the default settings provided in the MMDetection. The symbol "*" means the gradient of OPCL with respective to IoU is computed during training.}
 \begin{tabular}{c|c|cccccc} 
 \hline
Head & Backbone & mAP & $\text{AP}_{50}$ &$\text{AP}_{60}$ & $\text{AP}_{70}$ &$\text{AP}_{80}$ & $\text{AP}_{90}$\\
 \hline
baseline & ResNet-50-FPN & 79.8 & 95.2 & 93.6 & 90.5 & 84.5 & 63.7\\
baseline & ResNet-101-FPN & 81.6 & 95.7 & 94.5 & 91.8 & 86.1 & 67.4\\
baseline & ResNeXt-32x4d-101-FPN & 82.1 & 96.0 & 94.7 & 92.0 & 86.6 & 67.9\\
 \hline
OPCL & ResNet-50-FPN & 81.2 & 95.0 & 93.8 & 91.0 & 85.2 & 67.0\\
OPCL & ResNet-101-FPN & 82.9 & 96.0 & 94.8 & 92.2 & 87.2 & 70.1\\
OPCL & ResNeXt-32x4d-101-FPN & 83.6 & 96.2 & 95.2 & 92.6 & 87.6 & 71.3\\
 \hline
OPCL* & ResNet-50-FPN & 84.5 & 95.3 & 94.0 & 91.7 & 87.1 & 74.5\\
OPCL* & ResNet-101-FPN & 84.7 & 95.7 & 94.5 & 92.0 & 87.5 & 74.5\\
OPCL* & ResNeXt-32x4d-101-FPN & 86.5 & 96.4 & 95.3 & 93.4 & 89.4 & 77.8\\
 \hline
 \end{tabular}
  \label{table:carton}
 \end{center}
\end{table*}

\begin{table*}[t]
\begin{center}
\centering
   \caption{\footnotesize Experimental results on PASCAL VOC. All the models are trained on VOC2007 trainval and VOC2012 trainval and evaluated on VOC2007 test with the image scale of [600, 1000]. All the other settings are adopted as the same as the default settings provided in the MMDetection. The symbol "*" means the gradient of OPCL with respective to IoU is computed during training.}
 \begin{tabular}{c|c|cccccc} 
 \hline
Head & Backbone & mAP & $\text{AP}_{50}$ &$\text{AP}_{60}$ & $\text{AP}_{70}$ &$\text{AP}_{80}$ & $\text{AP}_{90}$\\
 \hline
baseline & ResNet-50-FPN & 51.4 & 79.1 & 74.6 & 64.0 & 45.4 & 15.9\\
baseline & ResNet-101-FPN & 55.1 & 81.1 & 77.2 & 67.5 & 50.4 & 20.1\\
baseline & ResNeXt-32x4d-101-FPN & 56.1 & 81.9 & 78.1 & 68.1 & 52.0 & 21.4\\
 \hline
OPCL & ResNet-50-FPN & 53.8 & 79.9 & 76.3 & 66.5 & 48.6 & 18.5\\
OPCL & ResNet-101-FPN & 56.1 & 81.4 & 77.9 & 68.9 & 52.0 & 20.8\\
OPCL & ResNeXt-32x4d-101-FPN & 57.9 & 82.6 & 79.4 & 70.0 & 54.9 & 23.6\\
 \hline
OPCL* & ResNet-50-FPN & 55.7 & 79.2 & 75.5 & 67.0 & 51.3 & 25.2\\
OPCL* & ResNet-101-FPN & 58.5 & 80.2 & 77.0 & 69.6 & 55.8 & 29.2\\
OPCL* & ResNeXt-32x4d-101-FPN & 59.6 & 78.3 & 70.7 & 70.7 & 57.0 & 30.7\\
 \hline
 \end{tabular}
  \label{table:voc}
 \end{center}
\end{table*}

\begin{table*}[t]
\begin{center}
\centering
   \caption{\footnotesize Experimental results on MS COCO. All the models are trained on COCO trainval and evaluated on COCO val-2017 with the image scale of [800, 1333]. All the other settings are adopted as the same as the default settings provided in the MMDetection. The symbol "*" means the gradient of OPCL with respective to IoU is computed during training.}
 \begin{tabular}{c|c|ccccccccc} 
 \hline
Head & Backbone & mAP & $\text{AP}_{50}$ &$\text{AP}_{60}$ & $\text{AP}_{70}$ &$\text{AP}_{80}$ & $\text{AP}_{90}$ & $\text{AP}_{S}$& $\text{AP}_{M}$& $\text{AP}_{L}$\\
 \hline
baseline & ResNet-18-FPN & 30.8 & 49.6 & 45.0 & 37.6 & 26.1 & 8.6 & 16.1 & 34.0 & 40.7\\
baseline & ResNet-50-FPN & 35.6 & 55.5 & 51.0 & 43.2 & 31.1 & 11.3 & 20.0 & 39.6 & 46.8\\
baseline & ResNet-101-FPN & 37.7 & 57.5 & 53.3 & 46.0 & 33.7 & 13.0 & 21.1 & 42.2 & 49.5\\
baseline & ResNeXt-32x4d-101-FPN & 39.0 & 59.4 &55.2  & 47.6 & 34.9 & 14.1  & 22.6 & 43.4 & 50.9\\
 \hline
OPCL & ResNet-18-FPN & 32.0 & 48.8 & 45.3 & 39.1 & 29.0 & 10.7  & 16.8 & 34.6 & 43.0\\
OPCL & ResNet-50-FPN & 36.5 & 54.8 & 51.1 & 44.6 & 33.1 & 13.1  & 20.5 & 40.2 &48.4\\
OPCL & ResNet-101-FPN & 38.7 & 57.2 & 53.5 & 47.2 & 35.9 & 14.6  & 21.6 & 43.0 &51.8\\
OPCL & ResNeXt-32x4d-101-FPN & 40.4 & 60.0 & 55.9 & 49.2 & 37.4 & 15.8 & 22.9 & 45.1 &53.3\\
 \hline
OPCL* & ResNet-18-FPN & 32.8 & 47.7 & 44.6 & 39.3 & 30.6 & 14.7  & 16.9 & 35.6 &44.6\\
OPCL* & ResNet-50-FPN & 37.4 & 54.1 & 50.4 & 44.8 & 35.0 & 16.8  & 20.8 & 41.4 &50.0\\
OPCL* & ResNet-101-FPN & 39.6 & 56.6 & 53.1 & 47.2 & 37.2 & 18.7  & 21.8 & 44.0 &53.7\\
OPCL* & ResNeXt-32x4d-101-FPN & 41.2 & 58.5 & 54.9 & 48.9 & 38.9 & 20.1 & 23.7 & 45.8 &54.8\\
 \hline
 \end{tabular}
  \label{table:coco}
 \end{center}
\end{table*}

\begin{table*}[t]
\begin{center}
\centering
   \caption{\footnotesize Main results of RetinaNet with all our proposed modules. "pretrain" means pretraining identity model on OSCD and fine-tuning on LSCD with the image scale of [600,1000]([800,1333]\dag). "1x" means the model is trained for total 12 epochs. In $\text{AP}_{S}$ item, these is on corresponding value because SCD are on small instance in the evaluation framework of MS COCO.}
 \begin{tabular}{c|c|c|ccccccc} 
 \hline
Detector & Backbone & Schedule & mAP & $\text{AP}_{50}$ &$\text{AP}_{75}$ & $\text{AP}_{S}$& $\text{AP}_{M}$& $\text{AP}_{L}$\\
 \hline
OPCL*+BGS &  ResNet-50-FPN & 1x & 85.2 & 96.0 & 90.1 & - & 48.2 & 85.3\\
OPCL*+BGS &  ResNet-101-FPN & 1x & 86.0 & 96.1 & 90.6 & - & 46.9 & 86.3\\
OPCL*+BGS &  ResNeXt-32x4d-101-FPN & 1x & 86.7 & 96.4 & 91.6 & - & 49.0 & 86.9 \\
OPCL*+BGS+pretrain & ResNet-50-FPN & 1x & 86.7 & 96.5 & 91.5 & - & 51.1 & 87.0\\
OPCL*+BGS+pretrain & ResNet-101-FPN & 1x & 87.5 & 96.5 & 92.4 & - & 50.0 & 87.7 \\
OPCL*+BGS+pretrain & ResNeXt-32x4d-101-FPN & 1x &87.5 & 96.5 & 92.2 & - & 48.1 & 87.8 \\
OPCL*+BGS+pretrain\dag & ResNet-50-FPN & 1x & 87.5 & 96.6 & 92.0 & - & 48.8 & 87.8\\
OPCL*+BGS+pretrain\dag & ResNet-101-FPN & 1x & 87.6 & 96.6 & 92.5 & - & 52.0 & 87.8\\
OPCL*+BGS+pretrain\dag & ResNeXt-32x4d-101-FPN & 1x & 89.0 & 97.2 & 93.7 & - & 55.5 & 89.2\\
 \hline  
 \end{tabular}
  \label{table:main}
 \end{center}
\end{table*}

\begin{figure}[!t]
\begin{center}
\includegraphics[width=\linewidth]{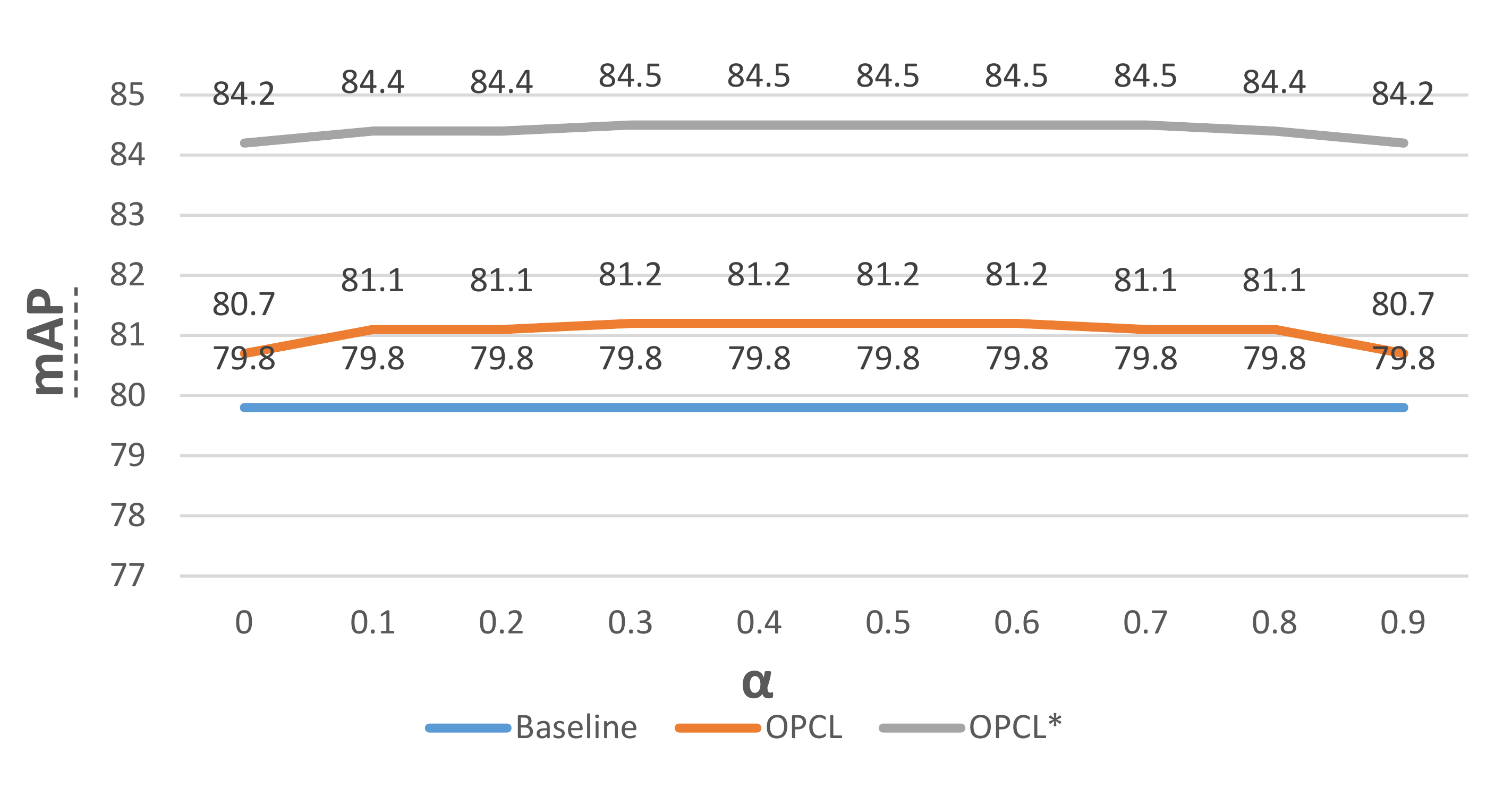}
\caption{The searching process about control factor $\alpha$ on OPCL. The symbol "*" means the gradient of OPCL with respective to IoU is computed during training}
\label{alpha}
\end{center}
\end{figure}

\begin{figure}[!t]
\begin{center}
\includegraphics[width=\linewidth]{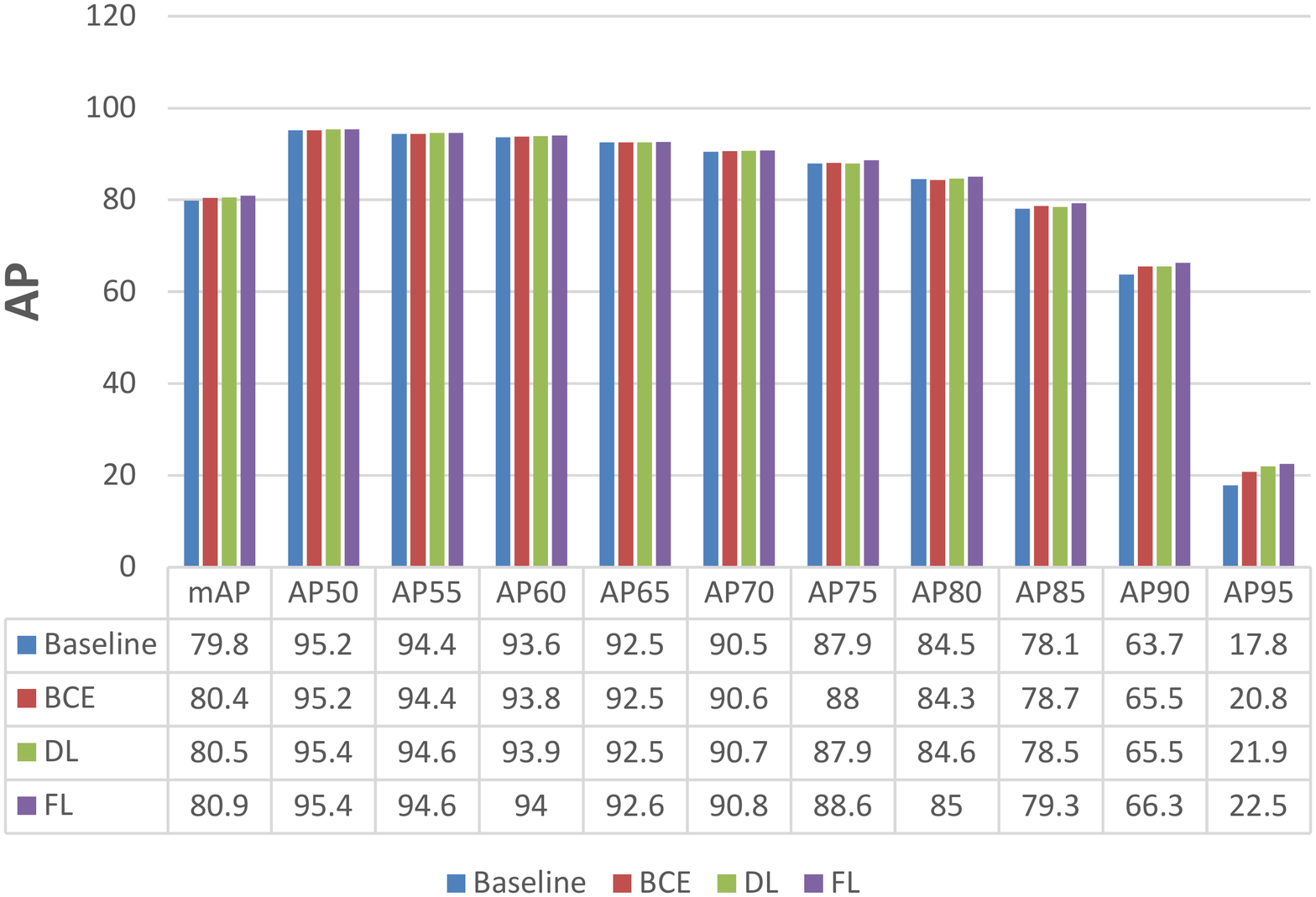}
\caption{Comparison of three type of losses to Boundary Guided Supervision module(BGS).}
\label{bgs_loss}
\end{center}
\end{figure}

\textbf{Ablation of the control factor of OPCL.} After predicting the offsets of classification score and localization accuracy, the important step is to integrate the offsets to classification score for inference. The control factor $\alpha$ is set from 0 to 1.0 to fine-tune the vanilla criterion which is fed into NMS procedure to implement better ranking. As Figure~\ref{alpha} shows, yellow and gray polyline describe the change of performance with respect to OPCL with closing and an opening gradient backpropagation. It indicates that the control factor $\alpha$ has an optimal interval while having an optimal value in PASCAL VOC and MS COCO.

\textbf{Generalization Experiments on PASCAL VOC 2007/2012 and MS COCO.} The behavior of OPCL in PASCAL VOC is shown in Table~\ref{table:voc}, when the gradient of OPCL with respective to IoU is not computed during training, OPCL equipped with different backbones improves AP by $1.0\% \sim 2.4\%$. When the gradient of OPCL with respective to IoU is opened during training, the AP is improved by $3.4\% \sim 4.5\%$ while the AP at higher IoU threshold(0.8 and 0.9) is improved by $9.1\% \sim 9.3\%$. The significant improvement about OPCL indicates that samples with worse localization accuracy are further improved quality. As shown in Table~\ref{table:coco}, the conclusions from the experimental results of the MS COCO dataset are consistent with those from the experimental results of the PASCAL VOC dataset, which demonstrates OPCL has generalization ability to other datasets and can be applied to different application scenes.

\subsection{Ablation Studies of BGS on SCD}
\textbf{Loss Function of BGS.} As Figure~\ref{bgs_loss} shows, binary cross entropy loss, dice loss\cite{dice} and focal loss\cite{Lin2017Focal} are applied to learn the boundary map. As described in section~\ref{section:4.3}, the boundary maps prediction suffers from an imbalance between positive and negative samples so that result shows better performance with respect to focal loss. It is worth noting that the BGS module mainly improves the accuracy of high AP ($\text{AP}_{85}$, $\text{AP}_{90}$, $\text{AP}_{95}$ improve $1.2\%$, $2.6\%$ and $4.7\%$ respectively), which indicates that the BGS plays a key role in improving localization capability.

The use of focal loss brings 2 hyper-parameter, where the focusing parameter $\gamma$ controls the strength of the modulating term while the balance factor $\alpha$ controls the ratio of positive and negative samples in the loss. We adopt a same strategy as RetinaNet to search the great hyper-parameters. With $\gamma = 0.5$ and $\alpha = 0.5$, FL yields a 1.1 AP improvement over the BCE loss and Dice loss.

\begin{table}[!tb]\footnotesize
\centering
   \caption{\footnotesize Ablation performance of the thick with respect to BGS. Thick means the boundary pixel thickness on the input image.}
 \begin{tabular}{c|cccccccc} 
 \hline
 Thick & 8 & 16 & 20 & 24 & 28 & 32 & 36 & 40\\
 \hline
 mAP & 80.6 & 80.6 & 80.6 & 80.8 & 80.7 & 80.7 & 80.8 & 80.9\\
 \hline
 \hline
 Thick & 44 & 48 & 52 & 56 & 60 & 64 & 72 & 96\\
 \hline
 mAP & 80.8 & 80.7 & 80.9 & 80.7 & 80.8 & 80.8 & 80.8 & 80.8\\
 \hline
 \end{tabular}
  \label{thick}
\end{table}

\textbf{Boundary Thick of BGS.} One of the most important design factors in BGS is how to design the boundary thick to supervise loss function. Boundary information for each instance is extracted from the corresponding mask. The stride of 0.5 unit boundary thickness is adopted to search for the optimal supervisory thickness. Because the BGS module is attached on ${P}_{3}$ which reduces the resolution of the original image by 8 times so that the actual stride of thickness is 4 pixels. As shown in Table~\ref{thick}, a series of isometric boundary thicknesses are searched, which surprisingly indicates that the mAP fluctuates less than 0.3 between pixel thickness 8 and 96. It shows that the boundary thickness has little effect on the performance. Theoretically, the minimum thickness should be more than 8, which prevents the boundary disappears due to down-sampling while too large one brings too much noisy information. However, deep works show great performance in anti-interference.

\subsection{Main Results of All Modules on SCD}
Table~\ref{table:main} shows the main results of RetineNet with OPCL, BGS and pretraining strategy which establish a strong baseline. As shown in Table~\ref{table:ablation}, all proposed  modules are added to RetinaNet to make a vanilla detector. BGS module supervises model to pay more attention to boundary information which strengthens the ability of localization and improves performance to $80.9\%$. The OPCL module further improves a significant performance to $85.2\%$ by bridging the gap of classification and localization quality. All modules improve baseline significant performance by $5.4\%$ without any increase in computation during inference. Finally, we transfer the model parameter from the same detector pre-trained on OSCD, which further boost $1.4\%$ in mAP.



\begin{table}[!tb]\footnotesize
\centering
   \caption{\footnotesize The performance of RetinaNet sequentially adding our proposed modules. All experiments adopt ResNet50 as backbone with the image scale of [600,1000].}
 \begin{tabular}{c|ccccccc} 
 \hline
Baseline &\checkmark &\checkmark &\checkmark &\checkmark &\checkmark  \\
BGS &  &\checkmark &\checkmark &\checkmark &\checkmark  \\
OPCL &  &  & \checkmark & \checkmark&\checkmark \\
OPCL* &  &  &   &\checkmark&\checkmark \\
pre-training &  &  & & &\checkmark \\
\hline
mAP & 79.8 & 80.9  & 82.1 & 85.2 & 86.7\\
 \hline
 \end{tabular}
  \label{table:ablation}
\end{table}

\section{Conclusion}
\label{section:6}

SCD is the first public dataset focusing on various carton layout scenes. SCD dataset consists of two subsets with up to 16,136 images which are collected from the Internet and diverse realistic scenarios. The number of instances in the images is distributed across sparse and dense scenes. The two subset of SCD(LSCD and OSCD) simulate industrial carton stacking scenes and various generalized scenes in reality respectively. OSCD downloaded from the Internet provides a priority in any carton-wise scenarios. SCD is annotated accurately at the pixel-wise level to be employed by detection and the instance segmentation using deep learning works. Based on SCD, several state-of-the-art models are evaluated. RetinaNet is selected to be plugged with OPCL and BGS which balances classification and localization quality, and guides the model to pay more attention to boundary information respectively. We believe that SCD can provide a powerful application for the carton work scene and a new challenge to the computer vision community. 

In the future, we plan to expand and refine various carton scenes and provide more labels for specific tasks, \ie, 3D detection, carton trademark detection, one-shot Detection, transfer learning and data augmentation.

{\small
\bibliographystyle{ieee}
\bibliography{ref}
}

\end{document}